\documentclass{article}

\PassOptionsToPackage{numbers,compress}{natbib}
\usepackage[preprint]{neurips_2026}

\usepackage[utf8]{inputenc}
\usepackage[T1]{fontenc}
\usepackage{amsmath,amssymb,mathtools}
\usepackage{graphicx}
\usepackage{booktabs}
\usepackage{algorithm}
\usepackage{algpseudocode}
\usepackage{nicefrac}
\usepackage{microtype}
\usepackage{xcolor}
\usepackage{url}
\usepackage{hyperref}
\hypersetup{colorlinks=true,linkcolor=blue!45!black,citecolor=blue!45!black,urlcolor=blue!45!black}
\usepackage{array}
\usepackage{tabularx}

\newcolumntype{Y}{>{\raggedright\arraybackslash}X}
\newcolumntype{P}[1]{>{\raggedright\arraybackslash}p{#1}}
\newcommand{\assetid}[1]{\path{#1}}

\makeatletter
\providecommand*{\theHALG@line}{}
\renewcommand*{\theHALG@line}{\arabic{algorithm}.\arabic{ALG@line}}
\makeatother

\newcommand{\R}{\mathbb{R}}

\newcommand{\snrdb}{\mathrm{SNR}_{\mathrm{dB}}}

\title{DiBA: Diagonal and Binary Matrix Approximation for Neural Network Weight Compression}

\author{Nobutaka Ono\thanks{Tokyo Metropolitan University, 6--6 Asahigaoka, Hino-shi, Tokyo 191--0065, Japan. \texttt{onono@tmu.ac.jp}}}

\begin{document}

\maketitle

\begin{abstract}
  In this paper, we propose DiBA (Diagonal and Binary Matrix Approximation), a compact matrix factorization for neural network weight compression. Many components of modern networks, including linear layers, $1\times1$ convolutions, attention projections, and embedding layers, have dense matrix weights. DiBA approximates $A\in\mathbb{R}^{m\times n}$ by $\widehat A=D_1B_1D_2B_2D_3$, where $D_1,D_2,D_3$ are diagonal matrices and $B_1,B_2$ are $0/1$ binary matrices. The intermediate dimension $k$ controls the trade-off between theoretical storage and approximation accuracy. For matrix-vector products, DiBA decomposes dense multiplication into three element-wise scaling operations and two binary mixing operations, reducing the floating-point multiplication count from $mn$ to $m+k+n$. For optimization, we introduce DiBA-Greedy, an alternating solver that combines closed-form least-squares updates for the diagonal factors with exact one-bit improvement tests for the binary factors. We also introduce DiBARD (DiBA with Retuning only Diagonal factors), which replaces dense-matrix layers by DiBA factors, freezes the binary matrices, and retunes only the diagonal entries on downstream data. This preserves compact binary mixing without discrete search during adaptation.  On 40 dense weight matrices extracted from public pretrained models, DiBA-Greedy yields consistent SNR improvements as the theoretical storage ratio increases. After DiBA replacement in two component-replacement studies, DiBARD improves DistilBERT/WikiText masked-token accuracy from 0.4447 to 0.5210 and Speech Commands test accuracy for an Audio Spectrogram Transformer from 0.7684 to 0.9781 without reoptimizing the binary factors.
\end{abstract}

\section{Introduction}
Dense matrices remain one of the dominant storage and computation units in modern neural networks.  Linear layers, token embeddings, attention projections, and $1\times1$ convolutions all store learned transformations as dense weight matrices.  These matrices often account for a large fraction of model size, memory traffic, and matrix-vector or matrix-matrix arithmetic at inference time.  Compressing pretrained dense weights while preserving downstream accuracy is therefore central to model distribution, on-device inference, low-memory deployment, and task-specific adaptation.

A broad literature has studied compact neural representations.  Pruning removes unimportant connections, filters, or channels, starting from classical second-order criteria such as Optimal Brain Damage and Optimal Brain Surgeon and extending to modern fine-tuning and one-shot pruning methods for pretrained models \citep{lecun1989obd,hassibi1993obs,han2015connections,li2017filterpruning,liu2017networkslimming,frankle2019lottery,sanh2020movementpruning,frantar2023sparsegpt,sun2024wanda}.  Quantization replaces real-valued weights by a small set of discrete codes, including vector quantization and weight sharing, binary and ternary networks, integer-arithmetic inference, post-training quantization, and weight-only quantization for large language models \citep{gong2014vectorquant,chen2015hashednets,han2016deepcompression,courbariaux2015binaryconnect,hubara2016bnn,rastegari2016xnor,zhou2016dorefa,zhu2017ttq,zhou2017inq,jacob2018quantization,esser2020lsq,nagel2020adaround,li2021brecq,dettmers2022llmint8,xiao2023smoothquant,frantar2023gptq,lin2024awq}.  Low-rank and tensor decompositions exploit linear structure in matrices and convolution kernels \citep{denton2014linear,jaderberg2014lowrank,novikov2015tensorizing}, while knowledge distillation transfers the behavior of a larger teacher model to a compact student \citep{hinton2015distilling,romero2015fitnets}.  Together, these approaches show that pretrained dense weights contain substantial redundancy, but they expose different trade-offs between storage, arithmetic structure, optimization difficulty, and downstream recovery.

Among these families, quantization is especially practical on existing hardware, but its optimization problem is intrinsically discrete.  In the simplest case, each scalar weight is rounded to the nearest quantization level.  In realistic low-bit settings, however, one must simultaneously choose discrete assignments, scales and zero points, layer-, channel-, or block-wise bitwidths, and the reconstruction or task loss to optimize after quantization.  As a result, low-bit quantization is often treated as a nonconvex problem with discrete constraints, mixed-integer structure, or binary assignment variables \citep{leng2018admm,nagel2020adaround}.  Existing methods use straight-through estimators \citep{bengio2013ste,courbariaux2015binaryconnect,hubara2016bnn}, learned step sizes or clipping parameters \citep{esser2020lsq,li2020apot}, ADMM or proximal formulations \citep{leng2018admm,bai2019proxquant}, loss-aware objectives \citep{hou2018lossaware}, Hessian- or second-order approximations for rounding and mixed precision \citep{dong2019hawq,dong2020hawqv2,nagel2020adaround,li2021brecq,frantar2023gptq}, and scale transformations that handle activation outliers or salient channels \citep{xiao2023smoothquant,lin2024awq}.  These techniques are powerful, but the underlying optimization remains hard in the very low-bit regime because it combines discrete search, continuous relaxation, local reconstruction, and calibration-data dependence.

We take a different route.  Rather than assign each individual weight to one of a few quantization levels, we approximate an entire dense matrix by interleaving diagonal scaling with shared $0/1$ binary mixing.  For a target matrix $A\in\R^{m\times n}$, DiBA represents
\begin{equation}
  \widehat A = D_1 B_1 D_2 B_2 D_3,
  \label{eq:intro_diba}
\end{equation}
where $D_1,D_2,D_3$ are real diagonal matrices, $B_1\in\{0,1\}^{m\times k}$ and $B_2\in\{0,1\}^{k\times n}$ are binary matrices, and the intermediate dimension $k$ controls the storage-accuracy trade-off.  The binary factors can be stored, in the theoretical accounting used in this paper, at one bit per entry, while the real-valued degrees of freedom are concentrated in the three diagonal factors.

\begin{figure}[t]
  \centering
  \includegraphics[width=1\linewidth]{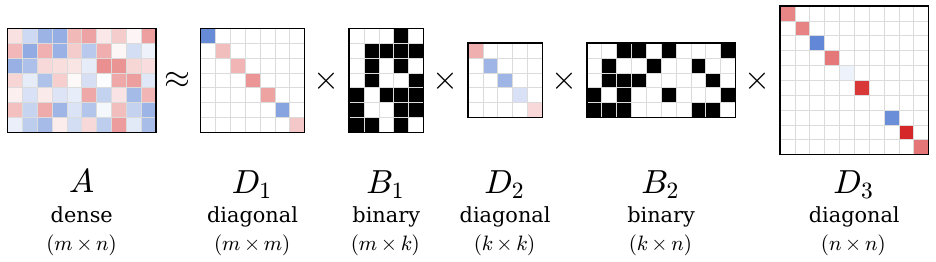}%
  \vspace{-2mm}
\caption{DiBA provides a structured approximation to a dense matrix using three real
diagonal factors and two \(0/1\) binary factors; \(k\) controls the
storage--accuracy trade-off.}
  \label{fig:diba_concept}
\end{figure}

This representation is useful for three reasons. First, the $0/1$ factors define shared selection-and-summation patterns rather than element-wise codebook entries; signs and magnitudes are carried by the diagonal scales. Second, although $\widehat A$ has rank at most $k$, DiBA is not an unconstrained real-valued low-rank factorization: its large factors are one-bit mixing patterns, and its real-valued degrees of freedom are diagonal. Third, the continuous and discrete parts admit a natural alternating optimization strategy: the diagonal factors can be refit by closed-form least-squares updates, while the binary factors can be improved by exact one-bit objective-difference tests in the corresponding binary subproblems.

We instantiate this idea with DiBA-Greedy, an alternating solver for matrix reconstruction.  Given fixed binary factors, DiBA-Greedy updates $D_1$, $D_2$, and $D_3$ by least squares.  Given fixed diagonal factors and the other binary factor, it evaluates whether flipping an entry of $B_1$ or $B_2$ decreases the squared Frobenius reconstruction error, accepting only flips whose exact objective change satisfies $\Delta < -\tau$ for a fixed flip tolerance $\tau$.  The method is not claimed to find a global optimum, but each accepted flip is an exact improvement for its local binary subproblem, and the continuous factors are repeatedly refit to the current binary structure.

We further introduce DiBARD, a diagonal-only retuning procedure for DiBA-replaced neural-network components. After replacing a dense neural-network weight matrix by DiBA factors, DiBARD freezes the binary matrices $B_1$ and $B_2$ and updates only the diagonal entries of $D_1$, $D_2$, and $D_3$ on downstream data.  This design avoids discrete search during task adaptation: the binary mixing structure remains fixed, and the recovery stage is a small continuous optimization over element-wise scales.  We do not claim measured speedups from bit-packed kernels in this work.  Our evaluation focuses on matrix approximation quality, theoretical storage ratio, and the extent to which diagonal-only retuning can recover downstream task performance.

Our contributions are as follows.  First, we formulate DiBA, a diagonal-and-binary factorization for approximating dense neural-network weight matrices, and derive its theoretical storage accounting.  Second, we propose DiBA-Greedy, which combines closed-form diagonal least-squares updates with exact one-bit greedy updates for binary factors.  Third, using 40 dense weight matrices extracted from public pretrained models, we evaluate the storage--SNR behavior of DiBA on real model weights. Fourth, we
propose DiBARD and show, in two component-replacement settings, DistilBERT/WikiText masked language modeling and AST/Speech Commands classification, that retuning only diagonal factors can recover downstream performance after DiBA replacement without reoptimizing the binary factors.

\section{Formulation}
\label{sec:formulation}

\subsection{Basic DiBA representation}
\label{subsec:basic_diba}
Let $A\in\R^{m\times n}$ be a dense target matrix.  DiBA approximates $A$ by a product of three real diagonal factors and two $0/1$ binary factors,
\begin{equation}
  \widehat A = D_1B_1D_2B_2D_3 .
  \label{eq:diba_basic}
\end{equation}
The factor sizes are
\begin{equation}
  D_1\in\mathcal{D}_m,\quad
  B_1\in\{0,1\}^{m\times k},\quad
  D_2\in\mathcal{D}_k,\quad
  B_2\in\{0,1\}^{k\times n},\quad
  D_3\in\mathcal{D}_n,
  \label{eq:diba_factor_sizes}
\end{equation}
where $\mathcal{D}_p$ denotes the set of real $p\times p$ diagonal matrices.  We write the diagonal degrees of freedom as vectors $d_1\in\R^m$, $d_2\in\R^k$, and $d_3\in\R^n$, defined by
\begin{equation}
  (d_1)_i=(D_1)_{ii},\qquad
  (d_2)_\ell=(D_2)_{\ell\ell},\qquad
  (d_3)_j=(D_3)_{jj} .
  \label{eq:diagonal_vectors}
\end{equation}
For a fixed intermediate dimension $k$, fitting DiBA amounts to solving
\begin{equation}
  \min_{D_1,D_2,D_3,B_1,B_2}
  \left\|A-D_1B_1D_2B_2D_3\right\|_F^2,
  \label{eq:diba_fit_problem}
\end{equation}
subject to the constraints in \eqref{eq:diba_factor_sizes}.  We refer to the product in \eqref{eq:diba_basic} as a DiBA approximation of $A$, and to $D_1,B_1,D_2,B_2,D_3$ as its DiBA factors.
We note the following.

\paragraph{Remark 1 (Binary rank-one superposition)}
Let $(B_1)_{:,\ell}$ denote the $\ell$th column of $B_1$ and $(B_2)_{\ell,:}$ the $\ell$th row of $B_2$.  The central factor expands as
\begin{equation}
  B_1D_2B_2
  =\sum_{\ell=1}^{k}(d_2)_\ell (B_1)_{:,\ell}(B_2)_{\ell,:} .
  \label{eq:binary_rank_one}
\end{equation}
Consequently,
\begin{equation}
  \widehat A_{ij}
  =(d_1)_i(d_3)_j
  \sum_{\ell=1}^{k}(d_2)_\ell (B_1)_{i\ell}(B_2)_{\ell j} .
  \label{eq:diba_entrywise}
\end{equation}
Thus $\widehat A$ is a superposition of $k$ scaled rank-one binary patterns with row- and column-wise diagonal corrections; in particular, $\operatorname{rank}(\widehat A)\le k$. We therefore view DiBA as a structured low-rank family whose distinction from conventional low-rank factorization lies in its $0/1$ shared mixing factors and bit-level storage accounting; unlike scalar quantization, entries are not rounded independently.

\paragraph{Remark 2 (Expressivity and non-identifiability)}
With a sufficiently large intermediate dimension, DiBA can represent any matrix $A\in\R^{m\times n}$ exactly.  One explicit construction sets $k=mn$ and associates each intermediate index $\ell$ with a single entry $(i,j)$.  Taking $D_1=I_m$, $D_3=I_n$, $(B_1)_{:,\ell}=e_i$, $(B_2)_{\ell,:}=e_j^\top$, and $(d_2)_\ell=A_{ij}$ represents every entry of $A$ by one binary rank-one pattern.\footnote{This construction is an expressivity argument only; it is not a compression regime.}  The factors are also not identifiable: rescalings among the diagonal factors and permutations of the intermediate channels can leave the product unchanged.

\subsection{Theoretical storage}
\label{subsec:storage}
Let $Q$ be the number of bits used to store one floating-point scalar in the dense matrix and in the diagonal entries.  A dense $Q$-bit representation of $A$ requires $S_{\mathrm{dense}}(Q)=Qmn\ \text{bits}.$ For DiBA, we count the entries of $B_1$ and $B_2$ at one bit per entry and the diagonal entries of $D_1,D_2,D_3$ at $Q$ bits per scalar.  Ignoring small metadata such as shapes, the theoretical storage is $S_{\mathrm{DiBA}}(k;Q)=k(m+n)+Q(m+k+n)\ \text{bits}.$
The theoretical storage ratio relative to a $Q$-bit dense matrix is therefore
\begin{equation}
  \rho_{\mathrm{DiBA}}(k;Q)
  =\frac{S_{\mathrm{DiBA}}(k;Q)}{S_{\mathrm{dense}}(Q)}
  =\frac{k(m+n)+Q(m+k+n)}{Qmn} .
  \label{eq:rho}
\end{equation}
A central motivation for DiBA is that $k$ directly tunes this storage-accuracy trade-off and can make the theoretical storage smaller than that of the dense representation.

\subsection{Matrix-vector product structure}
\label{subsec:mvp}
A standard dense matrix-vector product $y=Ax$ contains $mn$ floating-point multiplications per input vector when implemented as inner products.  With DiBA, the product can be evaluated from right to left as $y=D_1B_1D_2B_2D_3x.$
If multiplication by the $0/1$ matrices $B_1$ and $B_2$ is implemented as selection and summation rather than floating-point multiplication, then the floating-point multiplications come only from the three diagonal scalings.  This gives $m+k+n$ floating-point multiplications per input vector.  The remaining cost consists of additions, bit operations, and memory access for binary mixing, all of which depend on the implementation.  We therefore do not claim measured speedups or end-to-end runtime reductions in this paper; our experiments focus on approximation quality, theoretical storage, and downstream recovery by diagonal-only retuning.

\section{Optimization}
\label{sec:optimization}

\subsection{DiBA-Greedy}
\label{subsec:diba_greedy}
The fitting problem in \eqref{eq:diba_fit_problem} is nonconvex and includes binary variables.  We optimize it with \emph{DiBA-Greedy}, an alternating solver that separates the continuous diagonal refits from the discrete binary updates.  Given $B_1$ and $B_2$, the diagonal factors are updated by least-squares subproblems with closed-form solutions.  Given the current diagonal factors and the other binary factor, each binary update evaluates the exact change in squared Frobenius error caused by a one-bit flip and accepts only flips whose improvement is larger than a fixed tolerance.

At a high level, DiBA-Greedy initializes $D_1,D_2,D_3$ and random binary matrices $B_1,B_2$, refits the diagonal factors, and then alternates between updating $B_1$, refitting $D_1$, updating $B_2$, refitting $D_3$, and refitting $D_2$.  A flip is accepted only when its exact objective change satisfies $\Delta< -\tau$, where $\tau$ denotes a flip tolerance.  The solver is greedy and does not claim global optimality; for the ideal unregularized subproblems, each primitive update is an exact improvement or a least-squares refit.

\subsection{Closed-form diagonal updates}
\label{subsec:diagonal_updates}
Fix the binary factors $B_1$ and $B_2$.  The diagonal factors can then be updated as least-squares variables.  For the left diagonal factor, let $G=B_1D_2B_2D_3$.  The subproblem
\begin{equation}
  \min_{D_1\in\mathcal{D}_m}\left\|A-D_1G\right\|_F^2
  \label{eq:opt_d1_problem}
\end{equation}
decomposes over rows, giving
\begin{equation}
  (d_1)_i
  =\frac{\langle A_{i,:},G_{i,:}\rangle}{\left\|G_{i,:}\right\|_2^2} .
  \label{eq:update_d1}
\end{equation}
When the denominator is zero, the corresponding entry of $d_1$ does not affect the objective; in our implementation it is set to zero.  Similarly, with $G=D_1B_1D_2B_2$, the right diagonal factor is updated by
\begin{equation}
  (d_3)_j
  =\frac{\langle A_{:,j},G_{:,j}\rangle}{\left\|G_{:,j}\right\|_2^2} .
  \label{eq:update_d3}
\end{equation}

For the middle diagonal factor, let $G_L=D_1B_1$ and $G_R=B_2D_3$.  The subproblem is
\begin{equation}
  \min_{D_2\in\mathcal{D}_k}\left\|A-G_LD_2G_R\right\|_F^2 .
  \label{eq:opt_d2_problem}
\end{equation}
This is a least-squares problem over a linear combination of $k$ rank-one matrices.  The normal equations for $d_2$ are
\begin{equation}
  Fd_2=b,
  \qquad
  F=(G_L^\top G_L)\odot(G_RG_R^\top),
  \qquad
  b_r=(G_L)_{:,r}^\top A (G_R)_{r,:}^\top,
  \label{eq:update_d2}
\end{equation}
where $\odot$ denotes the Hadamard product.  Equation~\eqref{eq:update_d2} gives the unregularized least-squares update; for numerical stability, the implementation solves $(F+\lambda I)d_2=b$ with a small $\lambda$.

\subsection{Exact objective change for one-bit flips}
\label{subsec:flip_delta}
To handle updates of $B_1$ and $B_2^\top$ with the same formula, consider the left-diagonal binary subproblem
\begin{equation}
  \min_{B\in\{0,1\}^{p\times q}}
  \left\|\widetilde A-D_LBG_R\right\|_F^2 .
  \label{eq:leftdiag_problem}
\end{equation}
Here $\widetilde A\in\R^{p\times t}$, $D_L\in\mathcal{D}_p$, $B\in\{0,1\}^{p\times q}$, and $G_R\in\R^{q\times t}$.  We write the diagonal entries of $D_L$ as $d\in\R^p$.  Updating $B_1$ corresponds to $\widetilde A=A$, $B=B_1$, $D_L=D_1$, and $G_R=D_2B_2D_3$.  Updating $B_2$ is handled by applying the same subproblem to $B_2^\top$:
\begin{equation}
  \left\|\widetilde A-D_LBG_R\right\|_F^2
  =\left\|A^\top-D_3B_2^\top D_2B_1^\top D_1\right\|_F^2,
\end{equation}
with $\widetilde A=A^\top$, $B=B_2^\top$, $D_L=D_3$, and $G_R=(D_1B_1D_2)^\top$.

Now consider flipping a single entry $B_{ij}$.  Let $\delta_{ij}=1-2B_{ij}\in\{-1,+1\}$, so the flipped binary matrix is $B^{(ij)}=B+\delta_{ij}e_ie_j^\top$.  For fixed $D_L$ and $G_R$, the exact change in squared error is
\begin{equation}
\begin{aligned}
  \Delta_{ij}
  &:={}
  \left\|\widetilde A-D_LB^{(ij)}G_R\right\|_F^2
  -\left\|\widetilde A-D_LBG_R\right\|_F^2  \\
  &=2(1-2B_{ij})(Y_{ij}-Z_{ij})+h_ir_j .
\end{aligned}
  \label{eq:flip_delta_leftdiag}
\end{equation}
The quantities in \eqref{eq:flip_delta_leftdiag} are
\begin{equation}
\begin{gathered}
  H=G_RG_R^\top\in\R^{q\times q},\qquad
  h=d\odot d\in\R^p,
  \qquad
  r=\operatorname{diag}(H)\in\R^q,\\
  Y=(h\odot_{\mathrm{row}}B)H\in\R^{p\times q},\qquad
  Z=d\odot_{\mathrm{row}}(\widetilde A G_R^\top)\in\R^{p\times q}.
\end{gathered}
  \label{eq:leftdiag_workspace}
\end{equation}
Here $\odot$ is element-wise multiplication, and $u\odot_{\mathrm{row}}M$ denotes row-wise scaling of a matrix, i.e., the $i$th row of $M$ is multiplied by $u_i$.  DiBA-Greedy accepts only flips satisfying $\Delta_{ij}< -\tau$.

The subproblem in \eqref{eq:leftdiag_problem} decomposes by rows as
\begin{equation}
  \sum_{i=1}^{p}\left\|\widetilde A_{i,:}-d_iB_{i,:}G_R\right\|_2^2 .
  \label{eq:leftdiag_row_decomp}
\end{equation}
Therefore, flips chosen from distinct rows, with at most one flip per row, do not change each other's objective differences and can be applied simultaneously.  The objective change of such a batch is the sum of the selected $\Delta_{ij}$ values; since each accepted flip satisfies $\Delta_{ij}< -\tau$, the batch update also decreases the objective for fixed $D_L$ and $G_R$.

\section{Downstream Retuning}
\label{sec:dibard}

DiBA-Greedy fits DiBA factors by minimizing the squared reconstruction error of a pretrained dense weight matrix.  This objective is useful for obtaining a compact matrix approximation, but it is not the downstream loss of the full neural network.  Consequently, replacing a dense weight by its DiBA approximation can still leave a gap between theoretical storage reduction and task performance.  One possible remedy would be to reoptimize all DiBA factors on downstream data, but doing so would again introduce a difficult mixed continuous-discrete optimization problem because the binary factors $B_1$ and $B_2$ would have to be adapted inside a nonlinear network loss.  The exact one-bit improvement tests used by DiBA-Greedy rely on the quadratic structure of the reconstruction subproblem for $B_1$ or $B_2^\top$, which is not generally preserved for arbitrary downstream losses.

We therefore propose \emph{DiBARD} (DiBA with Retuning only Diagonal factors), a diagonal-only adaptation procedure for DiBA-replaced neural-network components.  For each dense weight matrix selected for replacement, we first compute DiBA factors with DiBA-Greedy.  We then freeze the binary factors and optimize only the diagonal entries of $D_1$, $D_2$, and $D_3$ on downstream training data.  The diagonal factors are treated as trainable element-wise scaling layers, while the binary factors remain fixed binary mixing layers, and gradients are backpropagated only to the diagonal entries.

This design has two advantages.  First, because the binary factors are not searched again on downstream data, the recovery stage becomes a small continuous optimization problem that can be handled by standard gradient methods.  Second, while DiBA-Greedy is tied to reconstruction of the original dense weight, DiBARD directly adjusts row scales, intermediate-channel scales, and column scales for the downstream task loss. This allows task-specific compensation using only $m+k+n$ trainable scalars per replaced $m\times n$ matrix; because retuning changes only diagonal values, the theoretical DiBA storage ratio is unchanged.

More general adaptation protocols are possible.  For example, one could fine-tune the diagonal factors together with biases, normalization parameters, or layers that were not replaced by DiBA.  In the experiments reported in this paper, however, we use the stricter diagonal-only protocol: for each DiBA-replaced component, we retune only the diagonal factors, while keeping the binary factors, associated bias terms, and all layers not replaced by DiBA fixed.

\section{Experiments}
\subsection{Experiment 1: intermediate-dimension sweep on real-model dense matrices}
\label{subsec:exp1_matrix_sweep}
Experiment~1 evaluates how DiBA-Greedy trades theoretical storage for matrix reconstruction quality on dense weight matrices extracted from public pretrained models.  We consider weights that can be represented as two-dimensional matrices, including embedding-module matrices, linear projections, and $1\times1$ convolutions.

\paragraph{Matrix collection and selection.}
We collected candidate matrices from Hugging Face NLP models and TorchVision vision models.  Each weight tensor was converted to a matrix whose rows correspond to output dimensions and whose columns correspond to input dimensions.  We canonicalized internally transposed projection weights, such as GPT-style projections, to the usual linear-layer orientation, and treated each $1\times1$ convolution as a $(C_{\mathrm{out}},C_{\mathrm{in}})$ matrix.  Shared or tied weights were counted only once.  We excluded matrices with either dimension at most 100, and matrices with more than $3{,}000{,}000$ entries.  This produced a pool of 885 unique matrices: 850 from Hugging Face/NLP models and 35 from TorchVision/vision models.  Across the pool, both row and column dimensions range from 112 to 3072, with median 512 for each dimension; the number of entries ranges from $16{,}384$ to $2{,}359{,}296$.

We grouped the pool into four functional categories, \texttt{attention\_related}, \texttt{ffn\_or\_projection}, \texttt{embedding}, and \texttt{conv1x1}, using layer names and model structure.  Here, \texttt{embedding} denotes matrices from embedding modules, including position-embedding tables and embedding-side projection or transformation weights; it is not restricted to token-embedding tables.  From each category we selected 10 matrices, for a total of 40 matrices.  Within each category, the selection was deterministic: candidates were ordered by matrix size, and representatives were chosen to cover a broad size range while avoiding excessive concentration on a single shape or model.  The selected set contains 26 Hugging Face/NLP matrices and 14 TorchVision/vision matrices.  Table~\ref{tab:exp1_selection} summarizes the candidate pool and the selected matrices.

\begin{table}[t]
\centering
\small
\caption{Candidate and selected matrices in Experiment 1. Categories are functional labels; the \texttt{embedding} category is defined in the text.}
\label{tab:exp1_selection}
\begin{tabular}{@{}lrrll@{}}
\toprule
Category & Candidates & Selected & Selected numel range & Source/modality \\
\midrule
attention-related & 424 & 10 & $16{,}384$--$1{,}769{,}472$ & hf/nlp:10 \\
FFN/projection & 419 & 10 & $65{,}536$--$2{,}359{,}296$ & hf/nlp:6, tv/vision:4 \\
embedding & 13 & 10 & $32{,}768$--$1{,}574{,}400$ & hf/nlp:10 \\
$1\times1$ conv & 29 & 10 & $32{,}768$--$512{,}000$ & tv/vision:10 \\
\bottomrule
\end{tabular}
\end{table}

\paragraph{Protocol and metric.}
For each selected matrix $A\in\R^{m\times n}$, we ran DiBA-Greedy at
intermediate dimensions $k\in\{8,16,32,64,128,256,512,1024\}$. We used $Q=16$ only for storage accounting, corresponding to FP16 storage for the dense matrix and for the DiBA diagonal entries; the DiBA-Greedy solver itself used float32 arithmetic.  We counted the binary factors at one bit per entry. The flip tolerance was $\tau=10^{-6}$ and the random seed was 0.  We fixed a storage-ratio cap $\rho_{\mathrm{DiBA}}(k;16)\leq 0.75$ before running the sweep and skipped configurations above this cap.  Of the planned $40\times8=320$ runs, 317 were completed; the only skipped runs were three small matrices at $k=1024$, and there were no failed runs.
We measure reconstruction quality by
$\snrdb = 10\log_{10} (\|A\|_F^2/\|A-\widehat A\|_F^2) .$

\begin{figure}[t]
\centering
\includegraphics[width=0.88\linewidth]{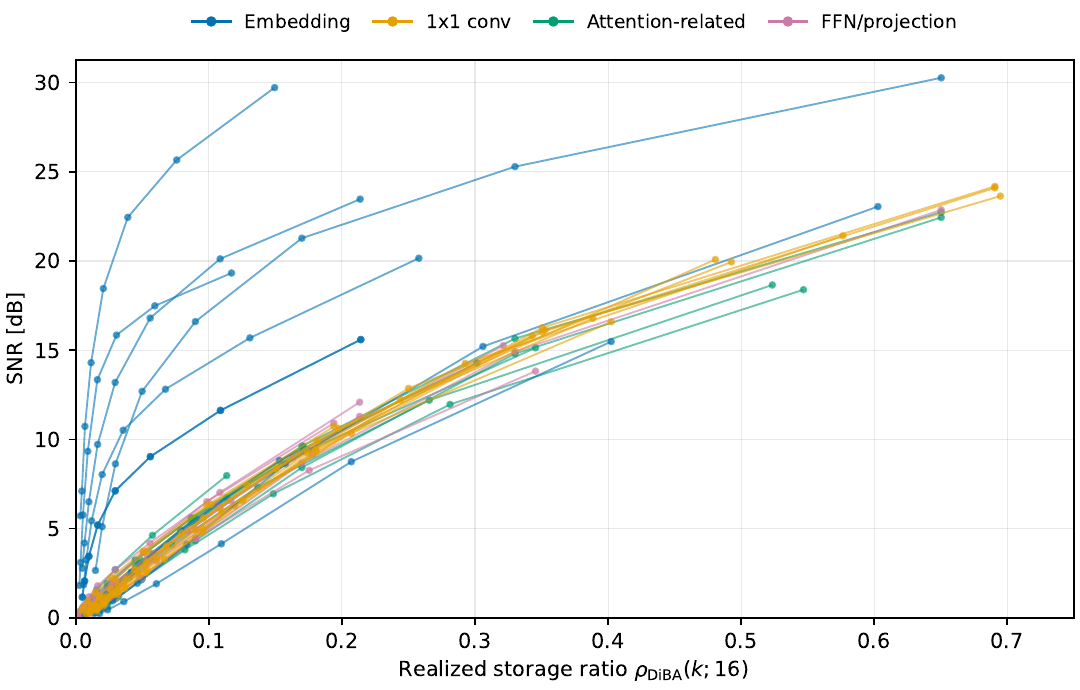}%
\caption{SNR versus realized DiBA storage ratio for the 40 selected matrices in Experiment~1. Each curve corresponds to one selected matrix, and markers correspond to the swept intermediate dimensions $k$. Colors indicate functional categories. The horizontal axis is the realized $\rho_{\mathrm{DiBA}}(k;16)$ computed from each matrix shape, not $k$ itself.}
\label{fig:exp1_per_matrix_snr_vs_rho}
\end{figure}

\paragraph{Results.}
Figure~\ref{fig:exp1_per_matrix_snr_vs_rho} plots the storage--SNR curve for each selected matrix.  We use the realized storage ratio $\rho_{\mathrm{DiBA}}(k;16)$ on the horizontal axis rather than $k$ itself, because the same intermediate dimension corresponds to different storage ratios for different matrix shapes.  Over the completed points, SNR is monotone nondecreasing in $k$ for all 40 matrix curves.  Averaged over matrices at each completed value of $k$, the mean SNR increases from 0.70~dB at $k=8$ to 16.35~dB at $k=1024$; the latter average is over the 37 matrices not removed by the storage-ratio cap.

The curves also suggest category-dependent behavior. Several matrices in the \texttt{embedding} category achieve higher SNR than other categories at comparable storage ratios, while many non-embedding curves from different model families and shapes occupy a similar storage--SNR band.  At $k=1024$, the mean SNRs over completed runs are 21.3~dB for the \texttt{embedding} category, 19.6~dB for $1\times1$ convolutions, 13.6~dB for attention-related matrices, and 11.5~dB for FFN/projection matrices. Since this experiment does not control for singular-value spectra or other matrix statistics, we treat these category differences as empirical tendencies rather than universal conclusions.

Overall, Experiment~1 shows that DiBA-Greedy consistently improves reconstruction quality as the theoretical storage budget increases on a diverse collection of real pretrained-model weight matrices, and that $\rho_{\mathrm{DiBA}}(k;Q)$ is a natural axis for comparing approximation quality across matrix shapes.

\subsection{Experiment 2-1: DiBARD on DistilBERT / WikiText MLM}
\label{subsec:exp2_1}
\paragraph{Setup.}
Experiment 2-1 evaluates whether diagonal-only retuning can recover masked language modeling (MLM) performance after replacing the tied token embedding / MLM output projection weight of \texttt{distilbert-base-uncased} by DiBA factors.  The replaced component is the tied vocabulary matrix with vocabulary size $V=30{,}522$ and hidden dimension $H=768$.  We first fit DiBA factors to the pretrained embedding matrix with DiBA-Greedy, using squared Frobenius reconstruction loss.  The intermediate dimension is $k=2048$ and the flip tolerance is $\tau=10^{-6}$.  During DiBARD retuning, we freeze $B_1$, $B_2$, the DistilBERT backbone, the MLM-head transform, and the output bias, and update only $d_1$, $d_2$, and $d_3$ using masked-token cross-entropy on the WikiText-103 train split.  We use AdamW with learning rate $3.0\times10^{-6}$, batch size 16, five epochs, and no explicit regularization.

We tokenize \texttt{wikitext-103-raw-v1}, concatenate the token stream, and split it into contiguous blocks of length 128, yielding 917{,}209 train blocks, 1{,}920 validation blocks, and 2{,}201 test blocks.  The train split is used for retuning, the validation split for checkpoint selection, and the test split for the final held-out evaluation.  We use MLM probability 0.15 with the standard BERT 80/10/10 masking rule.  Validation and test masks are fixed so that all methods are compared on the same examples and masked positions.  We compare against the original model, row-wise symmetric Int2 and Int4 post-training quantization (PTQ), Int4-scale-RT, DiBA before retuning, and DiBARD after retuning.  Int4-scale-RT keeps the Int4 integer codes fixed and retunes only a row-scale multiplier with the same masked-token cross-entropy objective.  Factor shapes, checkpoint selection, baseline details, and byte-level storage accounting are given in Appendix~\ref{app:exp2_1_details}.

\paragraph{Results.}
Table~\ref{tab:exp2_1} reports fixed-mask MLM results on 40,850 masked tokens. Before retuning, DiBA uses a smaller target storage ratio than Int4-PTQ and is 2.68 percentage points higher in accuracy. Retuning only $d_1,d_2,d_3$ raises DiBARD to 0.5210 accuracy, 7.63 points above DiBA before retuning and 1.00 point above Int4-scale-RT. Because retuning uses WikiText training data, DiBARD exceeding the original model should be interpreted as target-domain retuning, not zero-shot replacement.

\begin{table}[!htbp]
\centering
\small
\caption{Fixed-mask MLM results for Experiment~2-1. Target $\rho$ is for the replaced tied embedding/output weight only, excluding the output bias, and is not a full-model ratio.}
\label{tab:exp2_1}
\begin{tabular}{@{}lrrrrr@{}}
\toprule
Method & Target $\rho$ & \# retuned & Loss & PPL & Acc. \\
\midrule
Original & 1.0000 & 0 & 2.7651 & 15.88 & 0.4906 \\
Int2-PTQ & 0.0638 & 0 & 13.1310 & 504{,}318.34 & 0.0023 \\
Int4-PTQ & 0.1263 & 0 & 3.1223 & 22.70 & 0.4179 \\
Int4-scale-RT & 0.1263 & 30{,}522 & 2.6739 & 14.50 & 0.5110 \\
DiBA & 0.0869 & 0 & 3.0738 & 21.62 & 0.4447 \\
DiBARD & 0.0869 & 33{,}338 & 2.5869 & 13.29 & 0.5210 \\
\bottomrule
\end{tabular}
\end{table}

\subsection{Experiment 2-2: DiBARD on AST / Speech Commands}
\label{subsec:exp2_2}
\paragraph{Setup.}
Experiment 2-2 evaluates DiBARD on an Audio Spectrogram Transformer (AST) fine-tuned for Speech Commands classification.  The base model is \texttt{MIT/ast-\allowbreak{}finetuned-\allowbreak{}speech-\allowbreak{}commands-\allowbreak{}v2}, and the dataset is the \texttt{v0.02} split of \texttt{google/speech\_commands}.  The model is a 35-class classifier, while the dataset split additionally contains \texttt{\_silence\_}. We remove only the \texttt{\_silence\_} examples and map the remaining 35 labels to the model label IDs. The resulting split sizes are 84{,}843 train examples, 9{,}981 validation examples, and 4{,}482 test examples.

We replace the query, key, value, and attention-output dense projections in all 12 AST encoder layers, for a total of 48 Linear weights of size $768\times768$.  We do not replace the feed-forward layers, LayerNorm parameters, or classifier.  Each selected weight is factorized independently by DiBA-Greedy with squared Frobenius reconstruction loss and intermediate dimension $k=128$.  In DiBARD, all binary factors and all other pretrained model parameters are frozen; only $d_1$, $d_2$, and $d_3$ for the replaced layers are updated for one epoch on the train split with multiclass cross-entropy.  This gives 79{,}872 trainable scalar parameters.  We use Adam with learning rate $3.0\times10^{-6}$.  Baselines are the original model, row-wise symmetric Int2 and Int4 PTQ applied to the same 48 weights, DiBA before retuning, and DiBARD after retuning.  Storage accounting and the reconstruction-SNR profile of the replaced matrices are given in Appendix~\ref{app:exp2_2_details}.

\begin{figure}[t]
\centering
\begin{minipage}[t]{0.49\linewidth}
  \centering
  \includegraphics[width=\linewidth]{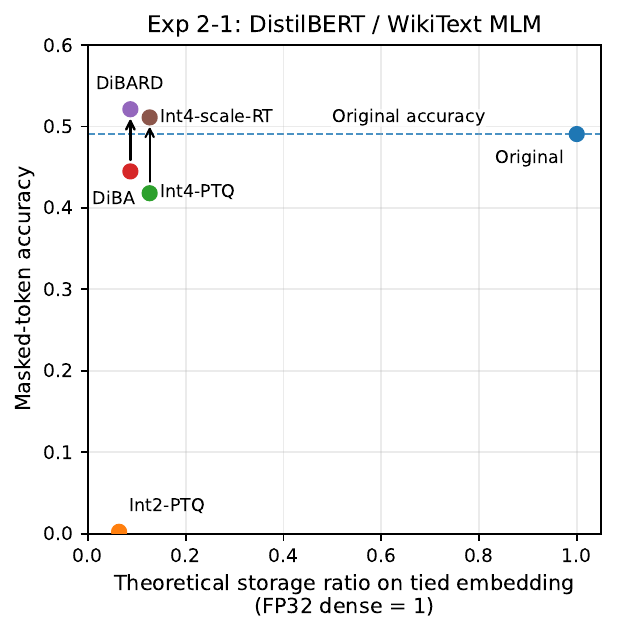}
  \vspace{-1mm}
  {\small (a) DistilBERT / WikiText fixed-mask MLM.}
\end{minipage}\hfill
\begin{minipage}[t]{0.49\linewidth}
  \centering
  \includegraphics[width=\linewidth]{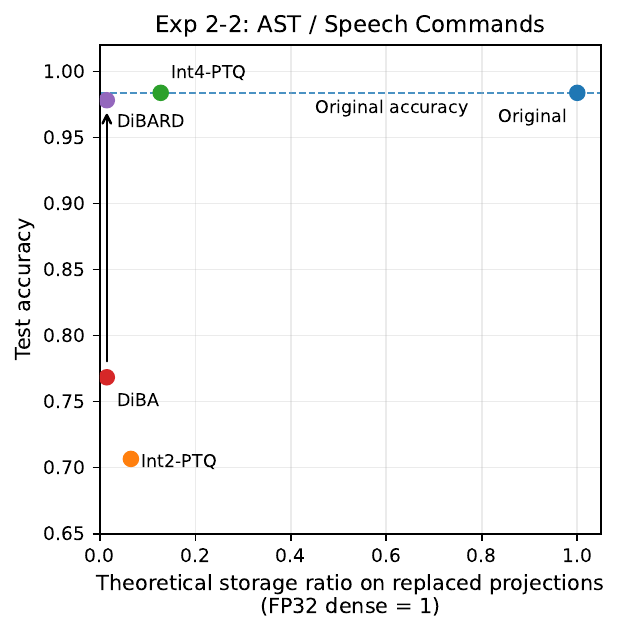}
  \vspace{-1mm}
  {\small (b) AST / Speech Commands classification.}
\end{minipage}
\caption{Held-out accuracy versus target-component theoretical storage ratio for Experiments~2-1 and 2-2. The horizontal axes are not full-model compression ratios; arrows denote scale-only retuning with fixed binary factors or fixed integer codes.}
\label{fig:exp2_downstream_storage_accuracy}
\end{figure}

\paragraph{Results.}
Table~\ref{tab:exp2_2} reports the final run. Int4-PTQ preserves the original accuracy on the same 48 projections, whereas DiBA before retuning reaches 0.7684 at target $\rho=0.0145$. DiBARD recovers to 0.9781 test accuracy, improving by 20.97 percentage points and recovering 97.4\% of the DiBA-induced gap. Figure~\ref{fig:exp2_downstream_storage_accuracy} summarizes the same fixed-discrete-structure recovery pattern across the two downstream experiments.

\begin{table}[!htbp]
\centering
\footnotesize
\caption{Speech Commands results for Experiment~2-2. Target $\rho$ and compression are theoretical values for the 48 replaced attention-projection weight+bias components only, with the corresponding FP32 dense components counted as 1; they are not full-model compression ratios.}
\label{tab:exp2_2}
\begin{tabular}{@{}lrrrrrr@{}}
\toprule
Method & Target $\rho$ & Comp. & Val. loss & Val. acc. & Test loss & Test acc. \\
\midrule
Original & 1.0000 & $1.00\times$ & 0.0889 & 0.9802 & 0.0712 & 0.9837 \\
Int2-PTQ & 0.0650 & $15.38\times$ & 1.3632 & 0.6787 & 1.1903 & 0.7066 \\
Int4-PTQ & 0.1274 & $7.85\times$ & 0.0900 & 0.9800 & 0.0692 & 0.9837 \\
DiBA & 0.0145 & $68.87\times$ & 1.2394 & 0.7282 & 1.0535 & 0.7684 \\
DiBARD & 0.0145 & $68.87\times$ & 0.1230 & 0.9726 & 0.0946 & 0.9781 \\
\bottomrule
\end{tabular}
\end{table}

\section{Conclusion}
\label{sec:conclusion}

DiBA approximates dense neural-network weights by interleaving diagonal scaling and shared $0/1$ binary mixing. Experiment~1 showed monotone nondecreasing SNR in the intermediate dimension $k$ over completed points as the theoretical storage budget increased on 40 real pretrained-model matrices, while Experiments~2-1 and 2-2 showed that DiBARD can recover downstream accuracy by retuning only diagonal entries with fixed binary factors. The main limitations are as follows: storage ratios are theoretical and assume bit-packed $B_1,B_2$; no measured memory, bit-packed-kernel, or inference-speed benchmark is provided; the storage ratios reported for Experiments~2-1 and 2-2 are computed with respect to the replaced target components only, not with respect to the full model;  DiBA-Greedy is a local greedy solver; and downstream evaluations use fixed splits with limited seeds and tasks. Future work includes packed implementations, broader layer/model coverage, and repeated-seed evaluation. Overall, DiBA offers a compact weight representation that combines fixed binary mixing with lightweight diagonal retuning.

\vspace{10mm}

\appendix

\section{DiBA-Greedy algorithm details}
\label{app:algorithm}

Algorithm~\ref{alg:diba_greedy} gives the outer alternating scheme used by DiBA-Greedy, and Algorithm~\ref{alg:row_greedy} gives the batch-row greedy step for the left-diagonal binary subproblem. Here $b$ is a row-batch size: each inner iteration of Algorithm~\ref{alg:row_greedy} flips at most one bit in each of up to $b$ selected rows, while one call to \textsc{RowGreedy} may contain multiple such inner iterations. At each inner iteration, the current best candidate difference $v_i$ is computed for every row; among rows satisfying $v_i<-\tau$, the method selects the rows with the largest improvements.

Assuming exact least-squares refits and exact flip-difference evaluations, every primitive update in DiBA-Greedy is monotone non-increasing for the reconstruction objective in Equation~\eqref{eq:diba_fit_problem}.  In the implementation, this monotonicity should be understood as a property of the idealized updates, because numerical roundoff and the regularization term used in Equation~\eqref{eq:update_d2} can slightly perturb the exact least-squares solution.

For the reported Experiment 1 matrix sweep, we used float32 arithmetic and
row-batch size $b=1024$.

\begin{algorithm}[b]
\caption{DiBA-Greedy}
\label{alg:diba_greedy}
\begin{algorithmic}[1]
\Require Matrix $A\in\mathbb{R}^{m\times n}$, intermediate dimension $k$,
flip tolerance $\tau\ge 0$, row-batch size $b\ge 1$.
\Ensure DiBA factors $D_1,B_1,D_2,B_2,D_3$.
\State Initialize $D_1,D_2,D_3$ and random binary matrices $B_1,B_2$.
\State Refit $D_1,D_2,D_3$ by Equations~\eqref{eq:update_d1}, \eqref{eq:update_d2}, and \eqref{eq:update_d3}.
\Repeat
  \State $(B_1,c_1)\leftarrow\textsc{RowGreedy}(A,d_1,D_2B_2D_3,B_1,\tau,b)$.
  \State Refit $D_1$ by Equation~\eqref{eq:update_d1}.
  \State $(B_2^\top,c_2)\leftarrow\textsc{RowGreedy}(A^\top,d_3,(D_1B_1D_2)^\top,B_2^\top,\tau,b)$.
  \State Refit $D_3$ by Equation~\eqref{eq:update_d3}.
  \State Refit $D_2$ by Equation~\eqref{eq:update_d2}.
\Until{$c_1+c_2=0$}
\State \Return $D_1,B_1,D_2,B_2,D_3$.
\end{algorithmic}
\end{algorithm}

\begin{algorithm}[t]
\caption{\textsc{RowGreedy}: batch-row greedy update for the left-diagonal
binary subproblem in Equation~\eqref{eq:leftdiag_problem}}
\label{alg:row_greedy}
\begin{algorithmic}[1]
\Require Target matrix $\widetilde A\in\mathbb{R}^{p\times t}$,
left scale $d\in\mathbb{R}^p$, right factor $G_R\in\mathbb{R}^{q\times t}$,
binary matrix $B\in\{0,1\}^{p\times q}$, flip tolerance $\tau\ge 0$,
row-batch size $b\ge 1$.
\Ensure Updated binary matrix $B$ and accepted flip count $c$.
\State $H\leftarrow G_RG_R^\top$, \quad $h\leftarrow d\odot d$, \quad $r\leftarrow\operatorname{diag}(H)$.
\State $Y\leftarrow(h\odot_{\mathrm{row}}B)H$.
\State $Z\leftarrow d\odot_{\mathrm{row}}(\widetilde A G_R^\top)$.
\ForAll{rows $i$}
  \State $j_i\leftarrow\arg\min_{1\le j\le q}\Delta_{ij}$, \quad $v_i\leftarrow\Delta_{i j_i}$ using Equation~\eqref{eq:flip_delta_leftdiag}.
\EndFor
\State $c\leftarrow0$.
\While{$\min_i v_i< -\tau$}
  \State $\mathcal E\leftarrow\{i\in\{1,\ldots,p\}: v_i< -\tau\}$.
  \State $\ell\leftarrow\min\{b,|\mathcal E|\}$.
  \State $\mathcal I\leftarrow$ the $\ell$ indices in $\mathcal E$ with the smallest $v_i$.
  \ForAll{$i\in\mathcal I$}
    \State $j\leftarrow j_i$, \quad $s\leftarrow1-2B_{ij}$.
    \State $B_{ij}\leftarrow1-B_{ij}$.
    \State $Y_{i,:}\leftarrow Y_{i,:}+s h_i H_{j,:}$.
    \State Recompute $j_i$ and $v_i$ for row $i$ using Equation~\eqref{eq:flip_delta_leftdiag}.
  \EndFor
  \State $c\leftarrow c+|\mathcal I|$.
\EndWhile
\State \Return $(B,c)$.
\end{algorithmic}
\end{algorithm}

\newpage

\section{Additional implementation details for Experiment 2-1}
\label{app:exp2_1_details}

This appendix gives only implementation details supplementing the protocol described in Experiment~2-1.  The same DiBA embedding module is used for both the input-side row lookup and the tied MLM vocabulary projection.  The resulting factor shapes are
$B_1\in\{0,1\}^{30522\times2048}$,
$B_2\in\{0,1\}^{2048\times768}$,
$d_1\in\R^{30522}$,
$d_2\in\R^{2048}$, and
$d_3\in\R^{768}$.

For DiBARD, we select the diagonal state with the lowest validation loss as the reporting checkpoint.  No parameter update or checkpoint selection is performed on the test split.  The frozen backbone is run in evaluation mode with dropout disabled, and gradient norms are clipped to 1.0.  Training masks are also generated deterministically by example index, while the validation and test masks are fixed as described in the main text to ensure identical examples and masked positions across methods. 

The Int4-scale-RT baseline fixes the integer codes obtained by row-wise symmetric Int4 PTQ and learns only a scale multiplier for each vocabulary row.  Its trainable parameters are $V=30{,}522$ log row-scale multipliers, and its loss is the same masked-token cross-entropy used by DiBARD.  We match the number of epochs and batch size to DiBARD and use learning rate $1.0\times10^{-4}$.

The Target $\rho$ values in Table~\ref{tab:exp2_1} are theoretical values for the tied weight matrix only, excluding the output bias; they are not full-model compression ratios.  The FP32 dense storage is counted as $30522\times768$ FP32 scalars, or 93{,}763{,}584 bytes.  Int2-PTQ uses 2-bit codes plus FP32 row scales, for 5{,}982{,}312 bytes.  Int4-PTQ and Int4-scale-RT use 4-bit codes plus FP32 row scales, for 11{,}842{,}536 bytes.  DiBA and DiBARD count $B_1$ and $B_2$ at one bit per entry and $d_1,d_2,d_3$ as FP32 scalars, for 8{,}143{,}592 bytes.  These counts correspond to target storage ratios 0.0638 for Int2, 0.1263 for Int4 and Int4-scale-RT, and 0.0869 for DiBA and DiBARD.

\section{Additional implementation details for Experiment 2-2}
\label{app:exp2_2_details}

This appendix gives only implementation details supplementing the protocol described in Experiment~2-2.  The Speech Commands v0.02 dataset labels consist of 35 words plus \texttt{\_silence\_}.  The AST classifier does not contain a \texttt{\_silence\_} class, so we remove only the \texttt{\_silence\_} examples.  The remaining 35 labels are mapped to model-side class IDs using the model's \texttt{id2label} metadata.  We therefore do not assume the dataset-side label-ID order, and all methods are evaluated with the same 35-class label-ID mapping.

For DiBA replacement, the original bias of each selected Linear layer is copied into the corresponding DiBALinear module and kept fixed during retuning.  Feed-forward layers, LayerNorm parameters, patch embedding, the classifier, and the biases of the selected Linear layers are not updated.  For runtime compatibility with dense matrix multiplication, an implementation may store the binary factors as floating-point buffers; however, the theoretical storage accounting in Table~\ref{tab:exp2_2} counts $B_1$ and $B_2$ at one bit per entry.

During DiBARD retuning, the frozen AST backbone is run in evaluation mode, and gradients are allowed to flow only to the diagonal parameters of the 48 DiBA replacements.  Weight decay and explicit diagonal regularization are set to zero, and gradient norms are clipped to 1.0.  Validation accuracy is evaluated after each epoch, and the state with the best validation accuracy is used for the final test evaluation.  Although the DiBA-before state is retained as a fallback checkpoint by the implementation, the one-epoch retuned state improves validation accuracy in the run reported here.

The storage ratios in Table~\ref{tab:exp2_2} are theoretical values for the 48 replaced attention-projection Linear weight+bias components only; they are not full-model compression ratios.  Bias terms are included in both the dense and compressed counts.  The FP32 dense storage of these components is 113{,}393{,}664 bytes.  Row-wise Int2-PTQ uses 7{,}372{,}800 bytes and row-wise Int4-PTQ uses 14{,}450{,}688 bytes, corresponding to $15.38\times$ and $7.85\times$ compression, respectively.  DiBA and DiBARD count $B_1$ and $B_2$ at one bit per entry and $D_1,D_2,D_3$ and the bias as FP32, for 1{,}646{,}592 bytes, corresponding to $68.87\times$ theoretical compression.

\begin{figure}[t]
\centering
\includegraphics[width=0.78\linewidth]{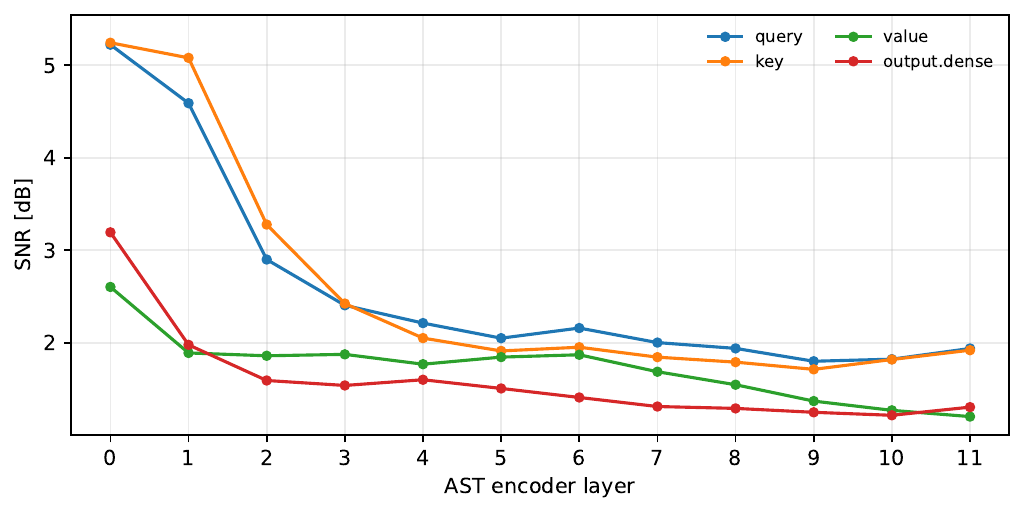}
\vspace{-2mm}
\caption{Per-layer reconstruction SNR for AST attention projections.}
\vspace{-2mm}
\label{fig:exp2_2_snr_by_layer}

\end{figure}

Across the 48 DiBA-factorized matrices, the factorization SNR has mean 2.13~dB, median 1.86~dB, minimum 1.20~dB, and maximum 5.24~dB.  By projection type, query and key are relatively higher at 2.59~dB and 2.58~dB on average, whereas value and attention-output dense projections average 1.73~dB and 1.60~dB.  Figure~\ref{fig:exp2_2_snr_by_layer} shows that the factorization quality varies
across projection types even at the same $768\times768$ shape and $k=128$:
query/key projections, especially in early layers, are reconstructed more
accurately than value and attention-output projections.

\end{document}